\documentclass{article}

\usepackage{PRIMEarxiv}

\usepackage[utf8]{inputenc} 
\usepackage[T1]{fontenc}    
\usepackage{hyperref}       
\usepackage{url}            
\usepackage{booktabs}       
\usepackage{amsfonts}       
\usepackage{nicefrac}       
\usepackage{microtype}      
\usepackage{lipsum}
\usepackage{fancyhdr}       
\usepackage{graphicx}       
\usepackage{tikz}
\usetikzlibrary{positioning}
\usepackage{amsmath}
\usepackage{float}

\title{From Kicking to Causality: Simulating Infant Agency Detection with a Robust Intrinsic Reward}

\author{
  Xia Xu, Jochen Triesch \\
  Frankfurt Institute for Advanced Studies \\
  Frankfurt am Main \\
  Germany \\
  \texttt{\{xiaxu, triesch\}@fias.uni-frankfurt.de}
}

\begin{document}
\maketitle

\begin{abstract}
A fundamental challenge in artificial intelligence and developmental psychology is understanding how an agent discovers its own causal efficacy. While human infants robustly solve this "contingency detection" problem within months, standard reinforcement learning agents remain brittle. Their reliance on correlation-based rewards renders them unable to distinguish self-generated effects from confounding environmental events, leading to a fragile sense of agency that fails to generalize to ecologically valid, noisy scenarios. To address this, we introduce the Causal Action Influence Score (CAIS), a novel, model-based intrinsic reward rooted in causal inference. CAIS formalizes agency detection by quantifying an action's influence as the 1-Wasserstein distance between two learned probability distributions: the distribution of sensory outcomes conditional on the agent's action, $p(h|a)$, and the baseline distribution of outcomes, $p(h)$. This divergence measure provides a robust reward signal that isolates the agent's specific causal impact from environmental noise. We test our approach in a simulated infant-mobile environment. While simple perceptual rewards suffice in a clean, deterministic setting, they fail completely when the mobile is subjected to confounding external forces. In stark contrast, the causality-based CAIS reward enables the agent to robustly filter out this noise, identify its influence, and learn the correct policy. Furthermore, the high-quality predictive model learned for CAIS proves to be a prerequisite for modeling more complex cognitive phenomena; when augmented with a surprise signal, our agent successfully reproduces the "extinction burst" observed when a learned contingency is unexpectedly removed. We conclude that explicitly modeling and inferring causality, rather than merely detecting correlation, is a crucial mechanism for an agent to develop a robust and generalizable sense of its own efficacy. This work provides a psychologically plausible computational model of a foundational aspect of cognitive development and offers a concrete framework for building more adaptive and capable autonomous systems that can function effectively in the unpredictable real world.
\end{abstract}


\section{Introduction}
A fundamental challenge spanning developmental psychology, neuroscience, and artificial intelligence is understanding how an agent discovers its own causal power—the capacity to act upon and wilfully influence its environment. Human infants solve this problem with remarkable efficiency. Within the first few months of life, they begin to detect the relationship between their actions and subsequent sensory events, a capacity known as contingency detection that forms the cornerstone of agency and social understanding \cite{bahrick_detection_1985, decasper_prenatal_1986, gergely_social_1996}. This natural proficiency stands in stark contrast to the fragility of many advanced artificial agents, which often require vast, hand-engineered reward functions and fail to adapt when environmental conditions deviate from their training data \cite{hafner_mastering_2024}. This gap highlights a search for the "unlearnable primitives" of intelligence—the core computational principles that bootstrap learning and development. We propose that a crucial such primitive is the ability to move beyond simple correlation and explicitly infer one's own causal influence on the world.\par

The classic mobile paradigm \cite{rovee_conjugate_1969, rovee-collier_topographical_1978, watson_reactions_1972}, in which an infant learns that kicking a specific limb activates an overhead mobile, provides a canonical testbed for exploring this initial discovery of agency. Replicating this seemingly simple feat in an artificial agent, however, exposes a deep flaw in prevailing computational models. Standard reinforcement learning (RL) approaches \cite{sutton_reinforcement_2018}, whether guided by external rewards or by common forms of intrinsic motivation such as novelty, prediction error, or surprise, are fundamentally based on detecting statistical correlation. While effective in clean, deterministic settings, these correlational methods are brittle in more ecologically valid scenarios where the sensory consequences of an agent's actions are obscured by confounding external events—in effect, environmental noise. An agent that merely correlates its kicking with the mobile's movement is computationally blind to the distinction between self-generated effects and random jiggles from an external force. This "causal ambiguity" leads to spurious conclusions and a fragile, unreliable sense of agency that fails to generalize—a core limitation of models that do not learn the underlying causal structure of their environment.\par

To overcome this limitation, we argue that a robust discovery of agency requires an agent to graduate from detecting correlation to explicitly inferring its causal influence. This work formalizes this principle by introducing a novel, model-based intrinsic reward called the Causal Action Influence Score (CAIS). CAIS operationalizes the psychological drive for competence and efficacy using the rigorous language of causal inference. Conceptually, it quantifies an action's influence by measuring the discrepancy between two probability distributions: the distribution of sensory outcomes conditional on the agent taking a specific action, $p(h|a)$, and the unconditional, baseline distribution of outcomes that occur naturally, $p(h)$. If an action has no effect, these distributions will be identical; if it has a significant causal impact, they will diverge. This divergence is measured using the 1-Wasserstein distance, a metric from optimal transport theory chosen for its unique ability to provide a robust, geometrically meaningful measure of distance even between complex, non-overlapping distributions, making it exceptionally well-suited for isolating a true causal signal from unstructured environmental noise. A high CAIS value provides a clean, intrinsic reward that drives the agent to learn and exercise its agency.\par

We test this framework using an embodied, simulated infant agent in the "MIMo-Mobile" environment, comparing four distinct intrinsic reward mechanisms across both deterministic and noisy conditions. Our results provide a clear and powerful demonstration of our central thesis.\par

First, in a simple, deterministic environment, we show that conventional, correlation-based rewards (e.g., those based on the magnitude of perceptual change) are sufficient to guide the agent to learn the correct contingency. This result explains the apparent success of such methods in simplified, laboratory-like settings.\par

Second, and most critically, when a significant confounding variable—a random external force—is applied to the mobile, these correlational rewards fail completely. The agent is unable to distinguish its own influence from the environmental noise. In stark contrast, the CAIS reward enables the agent to robustly filter out the confounding signal, correctly identify its causal influence, and learn the appropriate policy.\par

Third, we enhance the psychological plausibility of our model by showing that the robust world model learned for CAIS provides the necessary foundation for other cognitive phenomena. By augmenting the CAIS reward with a "surprise" signal that models expectation violation \cite{zaadnoordijk_movement_2020, alessandri_violation_1990}, our agent successfully reproduces the "extinction burst"—an intensified effort when a learned contingency is unexpectedly removed. An agent can only be meaningfully surprised if it has a strong, accurate expectation to violate; the high-quality causal model learned via CAIS provides exactly that, a synergy not achievable with brittle, correlation-based models.\par

This work makes the following contributions:
\begin{itemize}
    \item It introduces the \textbf{Causal Action Influence Score (CAIS)}, a novel, causality-based intrinsic reward that enables robust agency detection in noisy and confounded environments where traditional methods fail.
    \item It provides an empirical demonstration of the fundamental brittleness of correlation-based intrinsic rewards, showing they are artifacts of overly simplistic environments and not a viable path toward general agency.
    \item It presents a psychologically plausible computational model that simulates not only the initial learning of a sensorimotor contingency but also the extinction burst phenomenon, linking causal inference to expectation violation.
    \item It offers a concrete and robust framework for building more adaptive and autonomous agents that can discover their own capabilities and function effectively in the unpredictable real world.
\end{itemize}

\section{Simulation and Experimental Design}
\label{sec:sim-and-exp}

\begin{figure}[t]
    \begin{center}
        \begin{tikzpicture}
            \node(PanelA) {\includegraphics[width=0.24\textwidth]{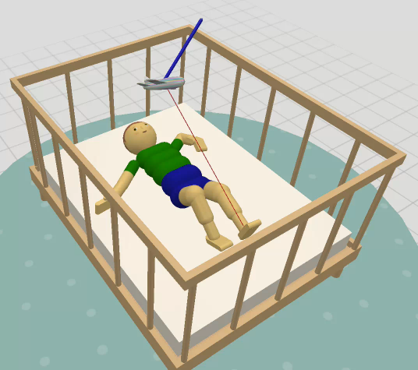}};
            \node[anchor=north east, xshift=-5pt, yshift=-5pt] at (PanelA.north east) {\textbf{A}};     
            
            \node[right=0cm of PanelA] (PanelB) {\includegraphics[width=0.24\textwidth]{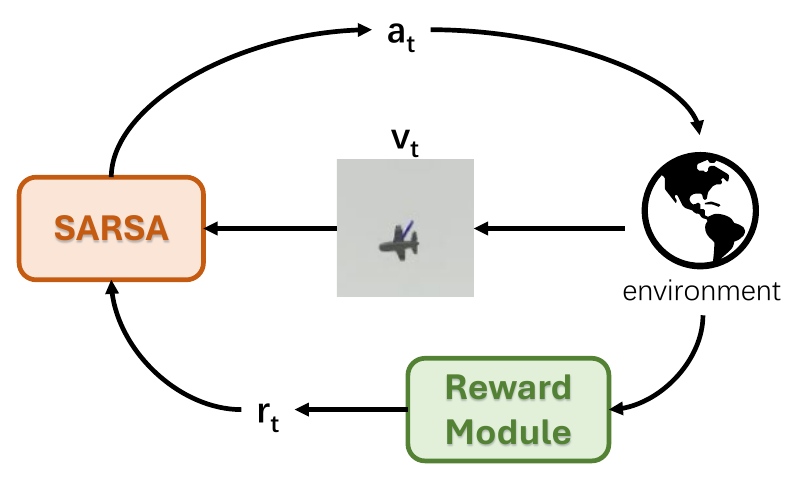}};
            \node[anchor=north east, xshift=-5pt] at (PanelB.north east) {\textbf{B}};
            
            \node[right=0cm of PanelB] (PanelC) {\includegraphics[width=0.24\textwidth]{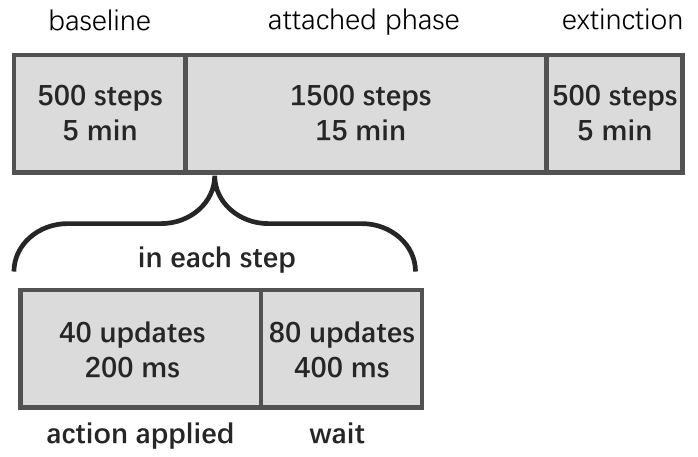}};
            \node[anchor=north east, xshift=-10pt, yshift=25pt] at (PanelC.south east) {\textbf{C}};
            
            \node[right=0cm of PanelC] (PanelD) {\includegraphics[width=0.24\textwidth]{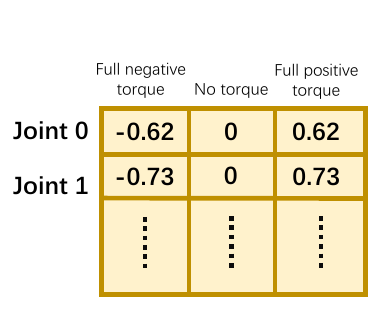}};
            \node[anchor=north east, xshift=-5pt, yshift=-5pt] at (PanelD.north east) {\textbf{D}};
        \end{tikzpicture}
    \end{center}
    
    \caption{Simulation setting. \textbf{Panel A}: MIMo-Mobile environment. The red rope connecting MIMo's left foot and the mobile is not visible to MIMo. \textbf{Panel B}: The paradigm of the interaction between the reinforcement learning-based agent and the environment. \textbf{Panel C}: Simulation duration. 120 environment updates are executed in each step. \textbf{Panel D}: The table illustrates the possible torque values; the agent's choice for each joint is a binary selection between $NO_TORQUE$ and $MOVE$, with the direction for MOVE being selected randomly.}
    \label{fig1}
\end{figure}
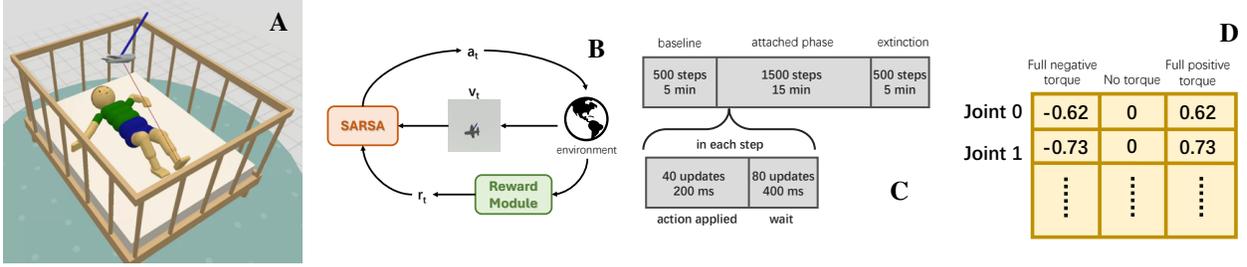

This section provides a comprehensive technical specification of the simulated environment, the embodied agent, the agent-environment interaction dynamics, and the phased experimental protocol. The methodology is designed to create a robust and reproducible testbed for investigating causal inference in an embodied agent, inspired by the canonical infant-mobile contingency paradigm.\par

\subsection{The MIMo-Mobile Physics Environment}
The agent's world is a virtual environment built upon MIMo (Multi-Modal Infant Model, \cite{mattern_mimo_2024}), an open-source simulation platform explicitly designed for research into early cognitive development. As detailed by \cite{mattern_mimo_2024}, MIMo provides a comprehensive model of an 18-month-old child, complete with detailed morphology and multiple sensory modalities. For this study, the agent is situated supine in a crib with a toy mobile suspended overhead (see Figure \ref{fig1}, Panel A), replicating the classic infant-mobile contingency paradigm.\par

All physical interactions within this environment are governed by the MuJoCo (Multi-Joint Dynamics with Contact) physics engine \cite{todorov_mujoco_2012}, which forms the core of the MIMo platform. The choice of MuJoCo is critical for reproducibility, as variations in physics simulators can lead to divergent results. MuJoCo is a standard in robotics research, selected for its high fidelity, computational efficiency, and stable handling of the complex, contact-rich dynamics that characterize the interactions in this paradigm. This foundation not only ensures the validity of the present study but also enhances the potential for future sim-to-real transfer of the learned policies.\par

\subsection{The Embodied Agent: Morphology, Perception, and Action}
The agent is an infant-like humanoid model named MIMo. Its morphology and sensorimotor capabilities are constrained to isolate the core learning problem.\par

\begin{itemize}
    \item \textbf{Morphology:} To focus the learning challenge on limb control, which is central to the mobile paradigm, MIMo's head and torso are fixed in place throughout the simulation. This mimics the typical posture of a young infant lying in a crib.
    \item \textbf{Perception:}  The agent's perception is multimodal, forming the state input for the reinforcement learning algorithm. For proprioception, the agent receives a 47-dimensional vector representing the current angles of all its joints, providing a complete and continuous sense of its body configuration. For vision, the agent perceives the world through a single, fixed virtual camera corresponding to its right eye, which generates a 128x128 RGB image at each time step. This raw visual input is processed by a pre-trained visual encoder (detailed in Section 3.2) into a compact 64-dimensional latent representation vector, denoted as $h_t$.
    \item \textbf{Actuation (Action Space):} The agent can apply torque to 12 specific joints in its four limbs (a full list is provided in the Appendix). The action space for each of these 12 joints is discrete (See Figure \ref{fig1}, Panel D). At each time step t, the reinforcement learning agent selects one of two possible actions for each joint $i$: $a_{i,t}\in\left\{NO\_TORQUE,MOVE\right\}$. If $NO\_TORQUE$ is selected, no torque is applied to that joint. If $MOVE$ is selected, a maximum torque is applied. The direction of this torque (positive or negative) is chosen randomly (50\%/50\% chance for each) for each $MOVE$ action. This action space design is a deliberate computational abstraction. It models the ballistic, all-or-nothing character of early infant movements, which lack the fine-grained control of mature motor skills. This simplification reduces the complexity of the exploration problem from a continuous space to a binary choice per joint. This can also be interpreted as a form of temporal abstraction, akin to concepts in Hierarchical Reinforcement Learning, where the agent selects a high-level motor primitive ("kick" or "rest") that unfolds over a fixed duration.
\end{itemize}

\subsection{Agent-Environment Interaction Loop}
The interaction between the agent and the environment is modelled as a discrete-time process (see Figure \ref{fig1}, Panel B). The simulation timing is standardized to ensure reproducibility.\par

Each agent step t corresponds to 600ms of simulated time. This interval is composed of 120 physics updates, executed at the simulator's rate of 200 Hz. The agent's chosen action vector $a_t$ is applied for the first 40 updates (200ms). This is followed by a 80-update (400ms) rest period during which no motor torques are applied, allowing the physical consequences of the action to unfold and the system to settle. At the conclusion of the 600ms interval, the resulting reward $r_t$ and the next state (composed of new proprioceptive and visual observations) are provided to the agent, which then informs the selection of action $a_{t+1}$.\par

\subsection{Experimental Protocol and Conditions}
The experiment unfolds over a total of 2500 agent steps, equivalent to approximately 25 minutes of simulated interaction time. This duration is divided into three distinct phases to test the agent's ability to learn, recognize, and react to the removal of a contingency (See Figure \ref{fig1}, Panel C).\par

\begin{itemize}
    \item \textbf{Phase 1: Baseline for 500 steps (from step 1 to 500; 5 min):} In this initial phase, the mobile is not attached to any of the agent's limbs and moves only in response to the simulation's physics (e.g., gravity, air resistance). This allows the agent to explore its own motor capabilities and provides a baseline measure of its behaviour and the mobile's dynamics.
    \item \textbf{Phase 2: Attached / Acquisition for 1500 steps (from step 501 to 2000; 15 min):} An invisible rope is introduced, connecting one of the agent's four limbs (a wrist or an ankle) to the mobile. Now, movements of the attached limb cause the mobile to move, establishing a direct contingency for the agent to discover.
    \item \textbf{Phase 3: Extinction for 500 steps (from step 2001 to 2500; 5 min):} The rope is removed, severing the learned contingency. The mobile no longer responds to the agent's actions. This phase is designed to test for phenomena such as the "extinction burst," where an agent initially intensifies its efforts when an expected outcome fails to occur.
\end{itemize}

To rigorously evaluate the robustness of the different reward mechanisms, we conduct the experiment under two distinct environmental conditions:\par
\begin{itemize}
    \item \textbf{Condition 1 (Free Mobile):} The standard setup described above, where the mobile's movement is determined solely by the agent's actions (during the attached phase) and passive physics.
    \item \textbf{Condition 2 (Noisy Mobile):} This condition is identical to the first, but with a significant confounding variable: a randomly-directed external force (40N) is applied to the mobile at every physics update. This creates a constant, noisy "jiggling" of the mobile that is independent of the agent's actions. This condition directly challenges the agent's ability to distinguish the sensory consequences of its own actions from spurious environmental events.
\end{itemize}

\section{Learning Framework and Intrinsic Rewards}
\label{sec:learning-and-rewards}

The agent's behaviour is driven by a reinforcement learning (RL) framework designed to discover and exploit its ability to influence the environment. This section details the learning algorithm, the state representation architecture, and the four distinct intrinsic reward mechanisms under investigation.\par

\subsection{Policy Learning with Expected SARSA}
The agent's policy is learned using the Expected SARSA algorithm \cite{sutton_reinforcement_2018}, an on-policy, temporal-difference (TD) method. The selection of this algorithm is deliberate and motivated by both technical and conceptual considerations.\par

From a technical perspective, Expected SARSA offers a significant advantage over standard SARSA in terms of update variance. Standard SARSA updates its value estimates based on the tuple $(s_t,a_t,r_t,s_{t+1},a_{t+1})$, meaning the update is subject to the randomness of the single sampled next action, $a_{t+1}$. In contrast, Expected SARSA computes the expected value of the next state by integrating over all possible next actions, weighted by their probabilities under the current policy, $\pi$. This removes the sampling variance associated with the selection of $a_{t+1}$, leading to more stable updates. This stability is particularly crucial in our setup, where the agent learns from an intrinsic reward signal (like CAIS) that is itself an online estimate and can be noisy, especially early in training. Reducing variance in the TD update rule prevents the compounding of noise from the reward signal and the action selection, making the overall learning process more robust.\par

From a conceptual standpoint, as an on-policy algorithm, Expected SARSA learns the value of the policy the agent is currently executing, including its exploratory actions. This is considered more psychologically plausible for modelling an infant's learning process, which is fundamentally grounded in its own direct, embodied experience, rather than learning an abstract optimal policy detached from its behaviour, as off-policy methods like Q-learning do.\par

The agent's action-value function, $Q(s,v)$, which estimates the expected return for taking an action given the proprioceptive state $s$ and visual state $v$, is defined as:

\begin{equation}
Q_{a_{t}^{i}}\left(s_{t},v_{t}\right)=\mathbb{E}\left[\sum_{k=0}^{\infty}\gamma^{k}R_{t+k+1}\mid a_{t}^{i}\right].
\label{eq:q_function}
\end{equation}

Action selection is governed by a Boltzmann (softmax) exploration strategy. The probability of selecting each of the two actions for a given joint is determined by their Q-values and a fixed temperature parameter, $\mathcal{T}$, which balances exploration and exploitation:

\begin{equation}
p\left(a_{t}^{i}=0\right)=\frac{e^{\frac{Q_{a_{t}^{i}=0}\left(s_{t},v_{t}\right)}{\mathcal{T}}}}{e^{\frac{Q_{a_{t}^{i}=0}\left(s_{t},v_{t}\right)}{\mathcal{T}}}+e^{\frac{Q_{a_{t}^{i}=1}\left(s_{t},v_{t}\right)}{\mathcal{T}}}}.
\label{eq:softmax}
\end{equation}

The Q-values are updated using the Expected SARSA temporal-difference error, $\delta_t$, which incorporates the immediate reward $r_t$ and the discounted expected value of the next state:

\begin{equation}
\delta_{t}=R_{t}+\gamma\left[\sum_{u\in\left\{ 0,1\right\} }p\left(a_{t+1}^{i}=u\right)Q_{a_{t+1}^{i}=u}\left(s_{t+1},v_{t+1}\right)\right]-Q_{a_{t}^{i}}\left(s_{t},v_{t}\right).
\label{eq:td-error}
\end{equation}

The action-value function is then updated using this error: $Q(s_{t},a_{t}^{i})\leftarrow Q(s_{t},a_{t}^{i})+\alpha\cdot\delta_{t}$, where $\alpha$ is the learning rate.

\subsection{State Representation and Visual Encoding}
The agent's state is a concatenation of the 47-dimensional proprioceptive vector $s_t$ (See Appendix) and the 64-dimensional latent visual representation $h_t$. The vector $h_t$ is produced by a visual encoder with a U-Net architecture\cite{ronneberger_u-net_2015}. This encoder is pre-trained on data from the MIMo-Mobile environment using a dual-objective function: a standard image reconstruction loss and an unsupervised temporal contrastive loss, specifically Contrastive Learning Through Time (CLTT \cite{schneider_contrastive_2021}, See Appendix \ref{appendix:encoder}). The CLTT loss encourages the model to learn smooth and informative representations of dynamics by pulling the latent codes of temporally adjacent video frames closer together. The weights of this visual encoder are frozen during the RL experiment, allowing it to function as a consistent feature extractor.

\subsection{A Taxonomy of Intrinsic Rewards}
\begin{figure}[t]
    \centering
        \includegraphics[width=0.8\textwidth]{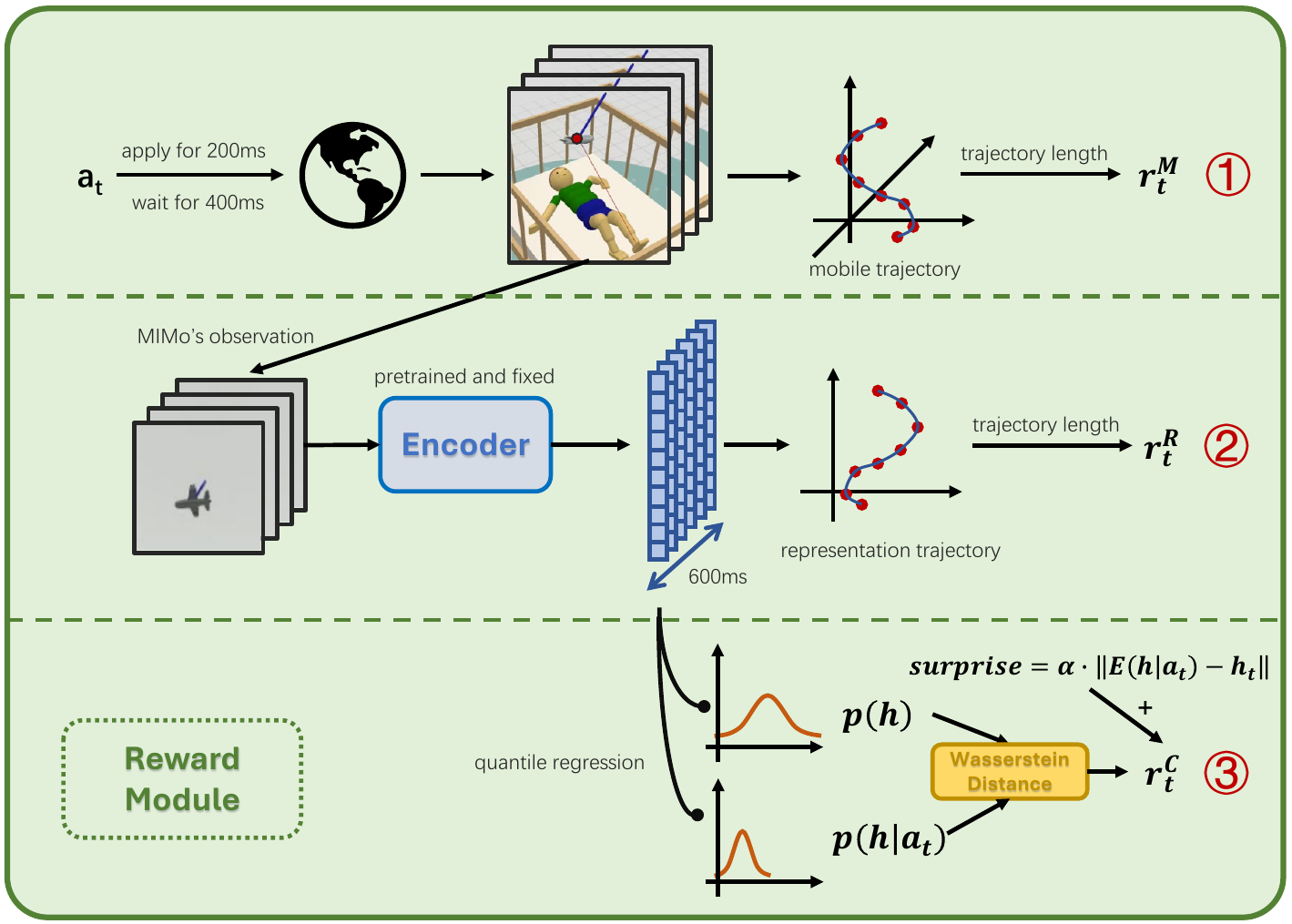}
    \caption{Different rewards. \textbf{Top}: The ground-truth mobile trajectory length (MTL). \textbf{Middle}: The representation trajectory length (RTL). \textbf{Bottom}: The Causal Action Influence Score (CAIS) module and the Surprise. The Surprise can also be used alone as a reward or combined with MTL or RTL reward.}
    \label{fig2}
\end{figure}

To investigate how an agent can discover its own efficacy, we compare four different intrinsic reward signals (See Figure \ref{fig2}). These signals are designed to represent a spectrum of computational strategies, from simple perceptual heuristics to explicit causal inference.\par

\begin{itemize}
    \item \textbf{Mobile Trajectory Length (MTL):} A ground-truth, non-plausible reward calculated as the total distance traveled by the mobile's center of mass in 3D space during one agent step. It serves as an "ideal observer" baseline, measuring the raw physical effect on the mobile.
    \item \textbf{Representation Trajectory Length (RTL):} A more biologically plausible perceptual reward. It is calculated as the Euclidean distance traversed by the latent visual vector $h_t$ in its 64-dimensional space over one agent step. This reward captures the magnitude of visual change perceived by the agent. It is important to note that because the latent space $h_t$ is learned via CLTT \cite{schneider_contrastive_2021}, which encourages temporal smoothness, this reward is not based on raw pixel changes but on a highly structured representation that already captures key aspects of the scene's dynamics.
    \item \textbf{Causal Action Influence Score (CAIS):} A model-based, causal-inferential reward. It is designed to explicitly quantify and isolate the agent's specific influence on the environment, filtering out confounding external factors. Its formulation is the central technical contribution of this work and is detailed in Section \ref{subsec:cais}.
    \item \textbf{Surprise:} A model-based reward reflecting the violation of the agent's expectations. It is calculated as the error between the agent's prediction of the sensory outcome and the actual outcome. It is tested both as a stand-alone reward and as an augmentation to the other three rewards to model phenomena like the extinction burst.
\end{itemize}

\subsection{The Causal Action Influence Score (CAIS)}
\label{subsec:cais}
The CAIS mechanism moves beyond simple correlation by formalizing contingency detection as a problem of causal inference. It provides a robust intrinsic reward by measuring the degree to which an agent's action systematically changes the distribution of its sensory inputs.\par

\subsubsection{Measuring Causal Influence with Wasserstein Distance}
Conceptually, CAIS quantifies an action's influence by measuring the discrepancy between two probability distributions:

\begin{itemize}
    \item The conditional distribution, $p(h|a)$, which represents the distribution of future visual states $h$ that are expected to occur after taking a specific action $a$.
    \item The unconditional distribution, $p(h)$, which represents the distribution of future visual states that occur naturally over time, reflecting the mixed consequences of the agent's entire behavioural policy. This serves as a baseline against which the specific effect of a single, filtered action can be compared.
\end{itemize}

If an action $a$ has no causal effect, then $p(h|a)$ will be identical to $p(h)$. If the action has a significant effect, these two distributions will diverge. The magnitude of this divergence is a measure of the action's causal influence.\par

This discrepancy is measured using the 1-Wasserstein distance, also known as the Earth Mover's Distance. The choice of this metric is critical. Unlike metrics such as Kullback-Leibler (KL) divergence, which can yield unstable or infinite values when comparing distributions with non-overlapping support, the Wasserstein distance provides a meaningful and smooth metric that reflects the underlying geometry of the outcome space. It intuitively measures the minimum "work" required to transform one distribution into the other, making it exceptionally well-suited for detecting a systematic shift in a distribution's location or shape, as would be expected from a causal action, even amidst significant environmental noise. The CAIS reward is thus defined as this distance:

\begin{equation}
    C\left(m\mid a\right)=W\left(p\left(m\mid a\right),p\left(m\right)\right) \;.
    \label{eq:causality_continuous}
\end{equation}

\subsubsection{Modeling Outcome Distributions with Quantile Regression}
To compute the Wasserstein distance, the agent must first learn to model the distributions $p(h|a)$ and $p(h)$. This work introduces a novel application of techniques from distributional reinforcement learning (DRL) for this purpose \cite{dabney_distributional_2018, dabney_distributional_2020}. While DRL algorithms typically use these techniques to model the distribution of future cumulative rewards, we repurpose them to model the distribution of immediate sensory outcomes.\par

Specifically, we use Quantile Regression to approximate the quantile function (the inverse Cumulative Distribution Function) for both the conditional and unconditional outcome distributions \cite{koenker_regression_1978}. Instead of learning the probabilities of a fixed set of outcomes, the agent learns the outcome values corresponding to a fixed set of quantiles. This non-parametric approach is flexible and can capture arbitrary distributions. A key advantage of this method is that the 1-Wasserstein distance between two distributions can be computed efficiently by integrating the absolute difference between their respective quantile functions \cite{vallender_calculation_1974}.\par

\begin{equation}
    W_{1}\left(p_{1},p_{2}\right)=\int_{0}^{1}|q_{1}\left(\tau\right)-q_{2}\left(\tau\right)|d\tau \;.
    \label{eq:quantile_w2}
\end{equation}

For each distribution, we model $N=49$ quantiles, $\tau\in{0.02,0.04,...,0.98}$. These models are trained online by minimizing the asymmetric quantile Huber loss. This loss function provides robustness to outliers while enabling the approximation of the full distribution. For a given quantile difference $u=y-q(\tau)$, where $y$ is the target value and $q(\tau)$ is the predicted value for quantile $\tau$, the loss is defined as:

\begin{equation}
    \mathcal{L}_{\tau,\kappa}(u)=|\tau-\mathbb{I}(u<0)|\cdot L_{\kappa}(u)
    \label{eq:quantile_loss}
\end{equation}
where $\mathbb{I}(\cdot)$ is the indicator function, which is 1 if its argument is true and 0 otherwise, and $L_{\kappa}(u)$ is the Huber loss with a threshold $\kappa$:
\begin{equation}
    L_{\kappa}\left(u\right)=\begin{cases}
    \frac{1}{2}u^{2}, & if\left|u\right|\leq\kappa\\
    \kappa\left(\left|u\right|-\frac{1}{2}\kappa\right), & {\rm  otherwise}
    \end{cases}.
    \label{eq:huber_loss}
\end{equation}
The $|\tau-\mathbb{I}(u<0)|$ term asymmetrically penalizes over- and underestimation, which is essential for learning the quantiles, while the Huber loss component provides stability by being quadratic for small errors and linear for large errors.\par

This approach-using DRL machinery to build a predictive world model for the purpose of causal inference—represents a key innovation of our framework.\par

\subsection{Modeling Expectation Violation for Extinction Bursts}
The "Surprise" reward is designed specifically to reproduce the psychological phenomenon of an extinction burst. The predictive models learned for the CAIS module, $p(h|a)$, naturally provide an expectation of the sensory outcome, $E[h|a]$. The surprise signal is defined as the prediction error—the distance between this expected outcome and the actual sensory outcome $h_{t+1}$ observed after the action is taken.\par

When a previously reliable contingency is removed (as in the extinction phase), the actual outcome will consistently violate the agent's learned expectation. This generates a large and persistent surprise signal, which, when used as a reward, drives the intensified exploratory behavior characteristic of an extinction burst.\par

\subsection{Implementation and Training Details}
The learning system is implemented using PyTorch. Both the Expected SARSA agent and the quantile regression models for CAIS/Surprise are trained using the AdamW optimizer \cite{loshchilov_decoupled_2019} with a fixed learning rate of 0.001.\par

To model the short-term, goal-directed behavior of infants in the mobile paradigm, the RL agent's discount factor is set to $\gamma=0.1$, making it highly myopic. The temperature for the Boltzmann exploration strategy is fixed at $\tau=0.3$ to ensure sufficient exploration throughout training.\par

To mimic a natural inclination towards rest and prevent the agent from constantly moving all its limbs, a small negative bias of -0.2 is added to the Q-value of the $MOVE$ action for every joint. This acts as a soft energy cost, encouraging sparser and more deliberate movements without overriding the influence of the learned intrinsic rewards.\par

\section{Results}
\label{sec:results}

This section presents the empirical evaluation of the four intrinsic reward mechanisms under two distinct environmental conditions. The analysis is structured to first establish a performance baseline in a simple, deterministic environment and then to rigorously test the robustness of each mechanism in a noisy, confounded environment. The findings demonstrate that while simple correlational rewards suffice in the former, only the proposed Causal Action Influence Score (CAIS) enables the agent to reliably discover its causal efficacy in the latter. Finally, by augmenting CAIS with a surprise-based signal, the model successfully reproduces the "extinction burst" phenomenon, lending it significant psychological plausibility.\par

\subsection{Performance in Standard vs. Noisy Environments}
\begin{figure}[t]
    \begin{center}
        \begin{tikzpicture}
            \node(PanelA) {\includegraphics[width=0.24\textwidth]{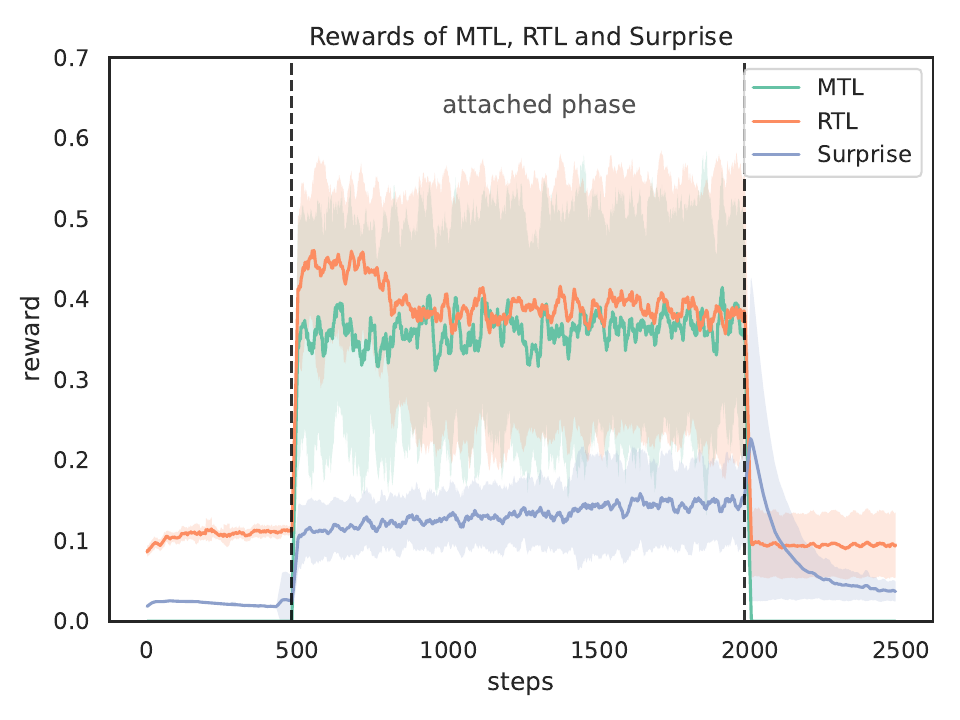}};
            \node[anchor=north west, xshift=20pt, yshift=-10pt] at (PanelA.north west) {\textbf{A}};     
            
            \node[right=0cm of PanelA] (PanelB) {\includegraphics[width=0.24\textwidth]{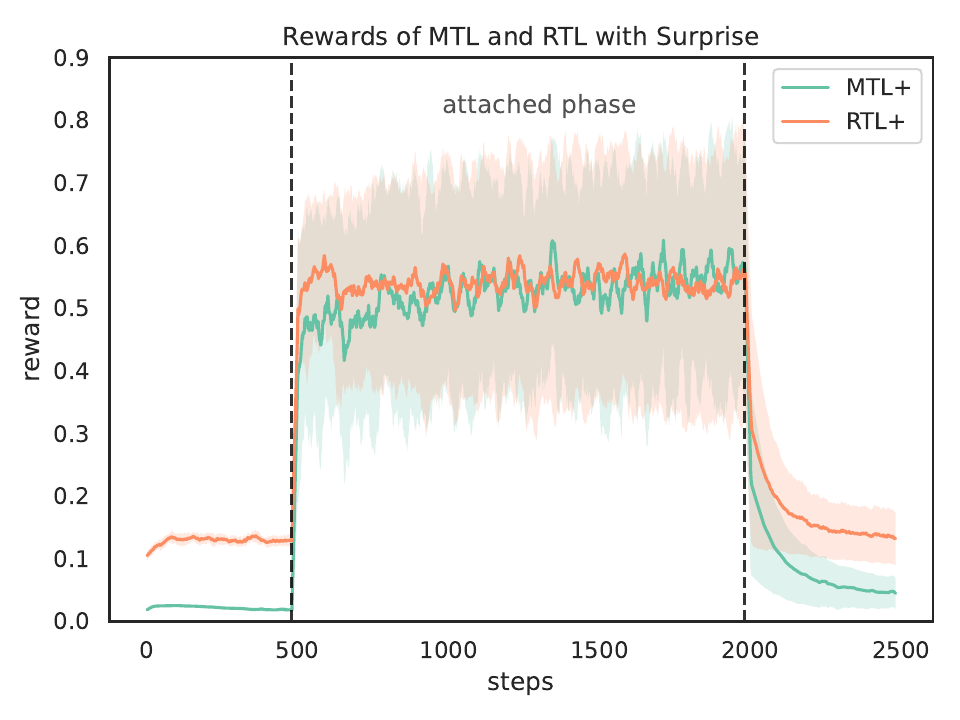}};
            \node[anchor=north west, xshift=20pt, yshift=-10pt] at (PanelB.north west) {\textbf{B}};
            
            \node[right=0cm of PanelB] (PanelC) {\includegraphics[width=0.24\textwidth]{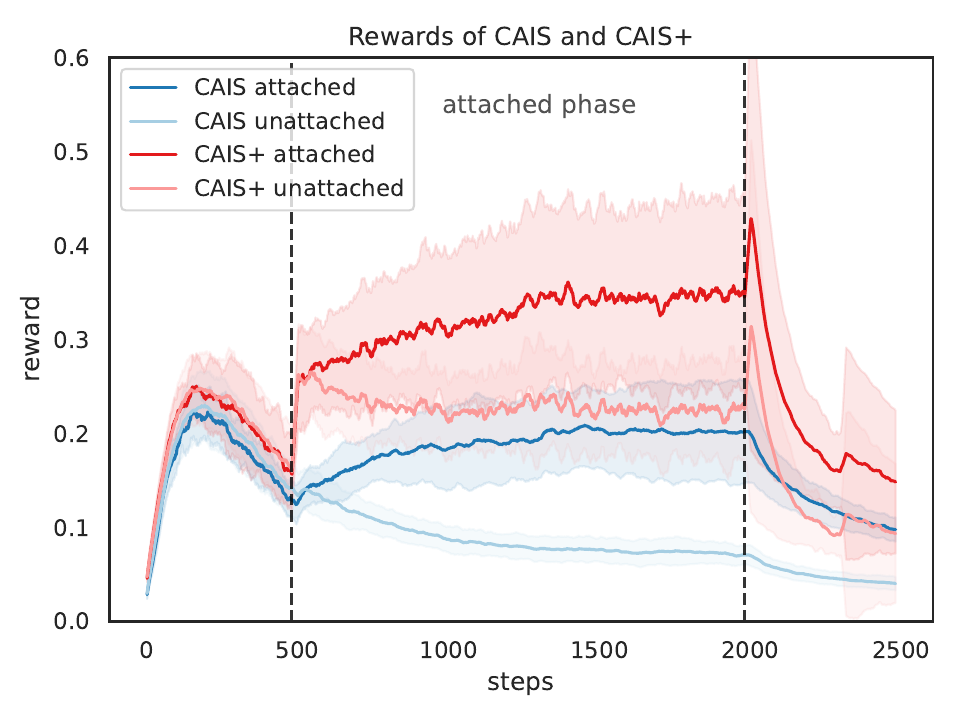}};
            \node[anchor=north west, xshift=-25pt, yshift=-10pt] at (PanelC.north east) {\textbf{C}};
            
            \node[right=0cm of PanelC] (PanelD) {\includegraphics[width=0.24\textwidth]{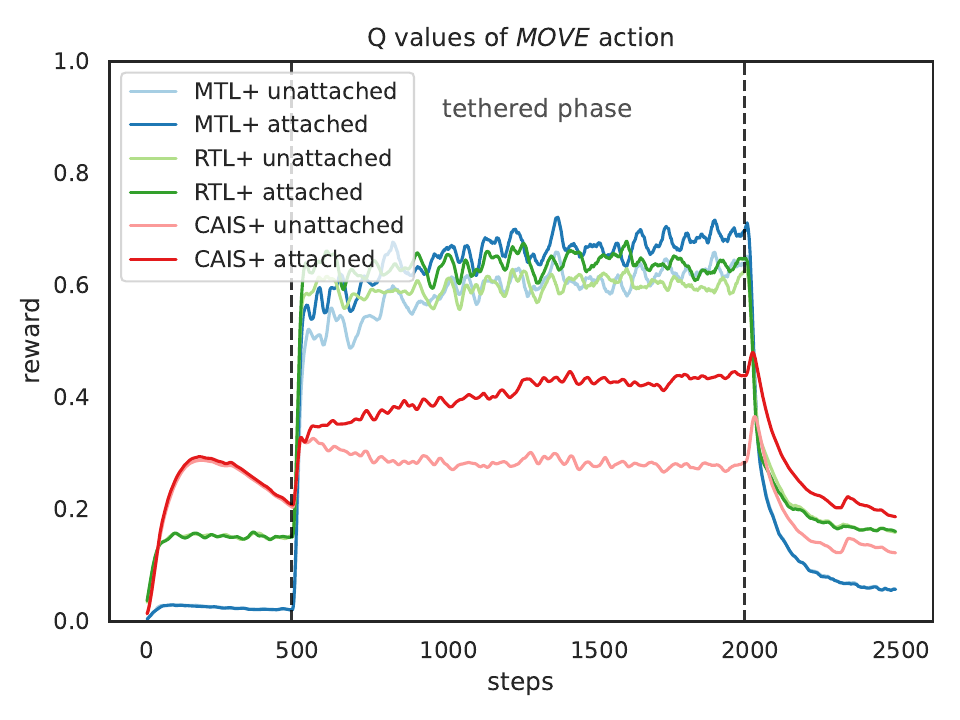}};
            \node[anchor=north west, xshift=-25pt, yshift=-10pt] at (PanelD.north east) {\textbf{D}};
        \end{tikzpicture}
    \end{center}
    
    \caption{Simulation results in the Free-Mobile condition. \textbf{Panel A}: MTL, RTL and Surprise reward. \textbf{Panel B}: MTL and RTL combined with Surprise. \textbf{Panel C}: CAIS and CAIS + Surprise reward. \textbf{Panel D}: Resultant Q values of the $MOVE$ action of different rewards combined with Surprise. Shaded area suggests 1 standard deviation.}
    \label{fig3}
\end{figure}
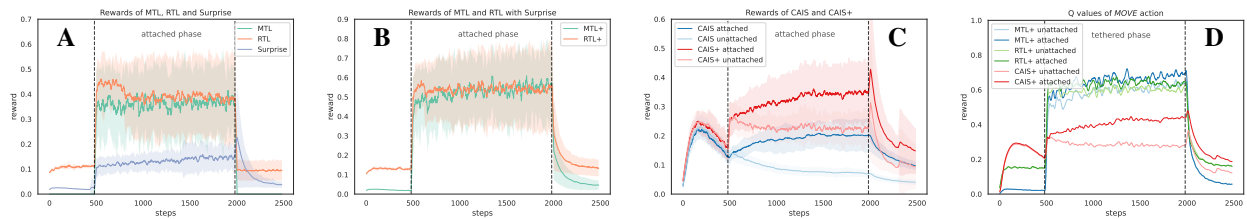

The initial set of experiments is conducted in the "Free-Mobile" condition, a standard, predictable environment where the mobile's movement is determined solely by the agent's actions and passive physics (See Figure \ref{fig3}). In this setting, all tested reward mechanisms proved capable of guiding the agent toward discovering the action-outcome contingency. The non-plausible, ground-truth Mobile Trajectory Length (MTL) and the perceptual Representation Trajectory Length (RTL) both showed a sharp and sustained increase during the "attached" phase, corresponding to the agent learning to move the contingent limb, and returned to baseline during the extinction phase, as shown in Figure \ref{fig3}, Panel A. This confirms that when the environment is free of confounding factors, the magnitude of physical or perceptual change serves as a reliable proxy for causal influence. The success of RTL specifically validates that the pre-trained visual encoder effectively captures the salient dynamics of the scene, such that the Euclidean distance in the latent space, $\left\|h_t-h_{t-1}\right\|$, is a faithful proxy for physical motion. Similarly, the Causal Action Influence Score (CAIS) reward for the limb attached to the mobile became significantly and consistently higher than for unattached limbs during the "attached" phase, as seen in Figure \ref{fig3}, Panel C. This validates that the CAIS mechanism correctly identifies that actions with the contingent limb lead to a systematic shift in the distribution of future sensory states. Ultimately, agents trained with RTL, MTL, or CAIS all successfully learned to increase the Q-value for the $MOVE$ action on the contingent limb, demonstrating that in a clean environment, all three signals provide a sufficient gradient for the agent to learn the correct policy (Figure \ref{fig3}, Panel D).\par

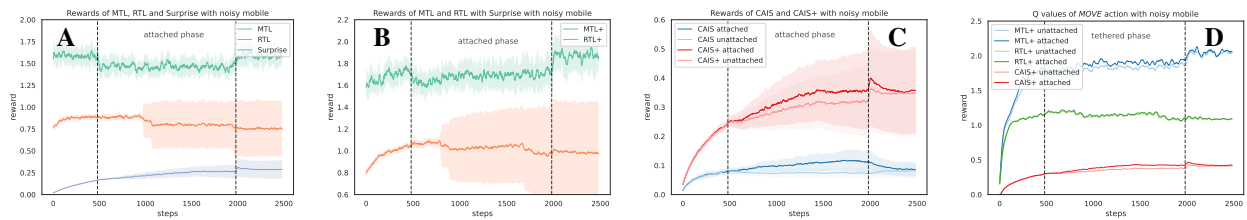
\begin{figure}[h]
    \begin{center}
        \begin{tikzpicture}
            \node(PanelA) {\includegraphics[width=0.24\textwidth]{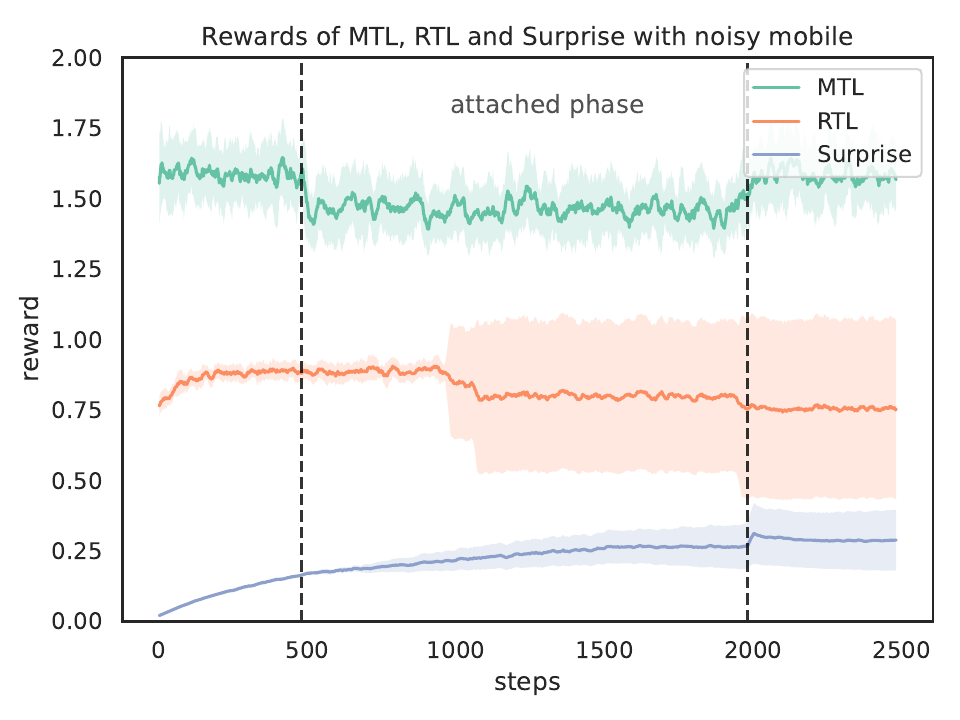}};
            \node[anchor=north west, xshift=20pt, yshift=-10pt] at (PanelA.north west) {\textbf{A}};    
            
            \node[right=0cm of PanelA] (PanelB) {\includegraphics[width=0.24\textwidth]{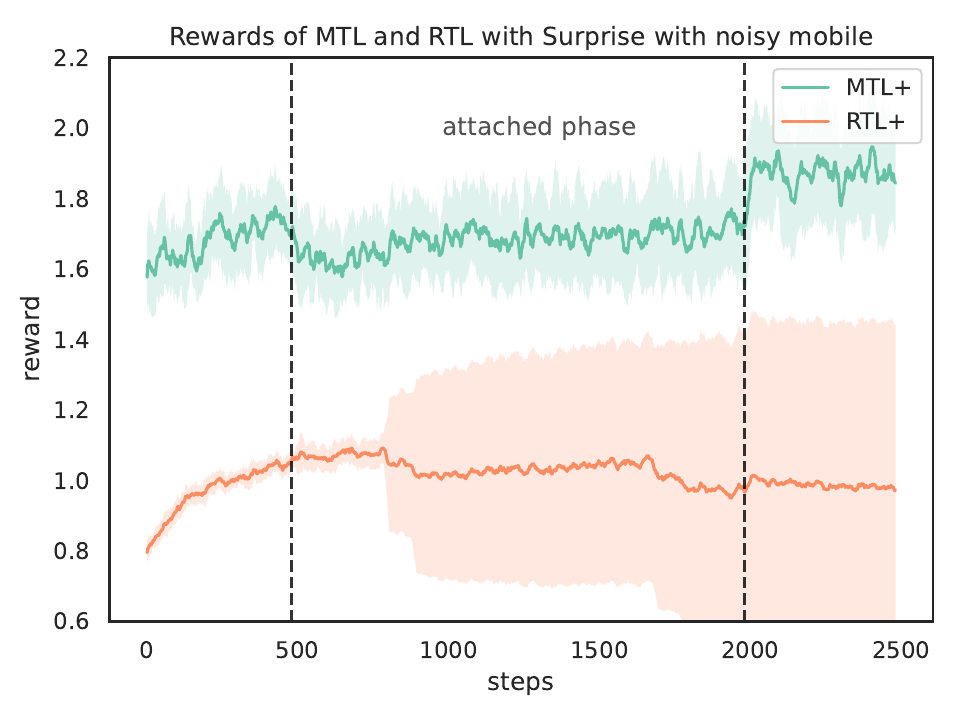}};
            \node[anchor=north west, xshift=20pt, yshift=-10pt] at (PanelB.north west) {\textbf{B}};
            
            \node[right=0cm of PanelB] (PanelC) {\includegraphics[width=0.24\textwidth]{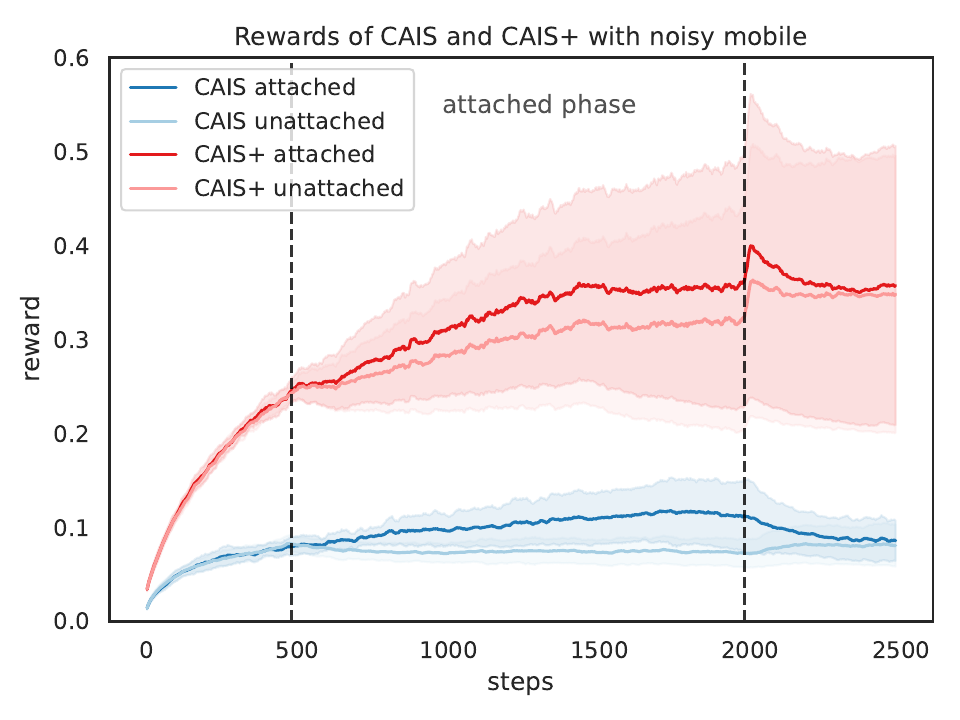}};
            \node[anchor=north west, xshift=-25pt, yshift=-10pt] at (PanelC.north east) {\textbf{C}};
            
            \node[right=0cm of PanelC] (PanelD) {\includegraphics[width=0.24\textwidth]{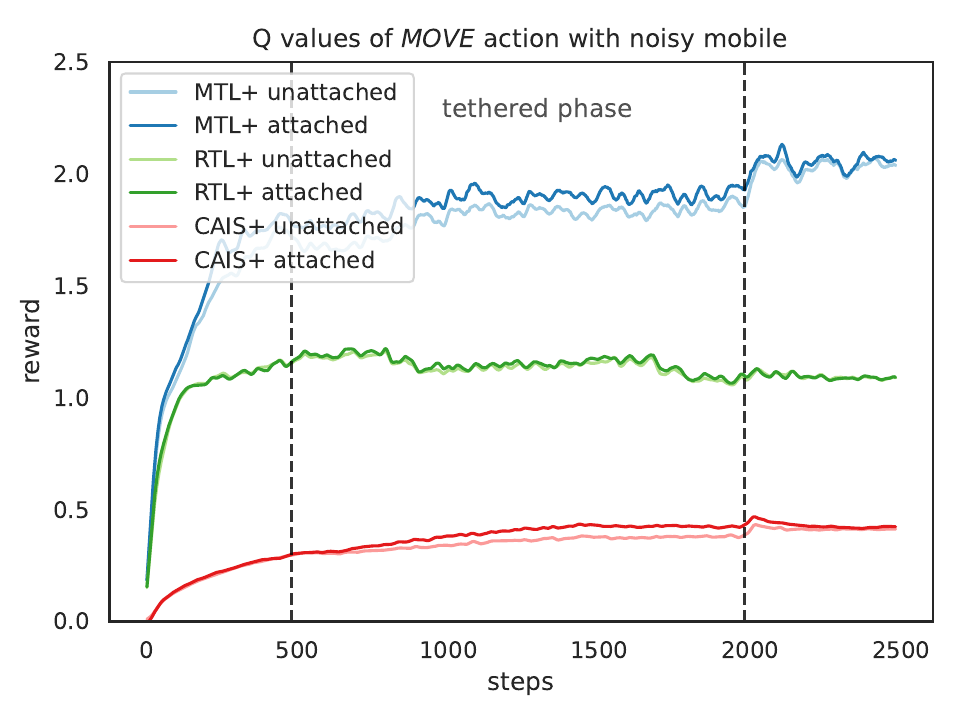}};
            \node[anchor=north west, xshift=-25pt, yshift=-10pt] at (PanelD.north east) {\textbf{D}};
        \end{tikzpicture}
    \end{center}
    
    \caption{Simulation results in the Noisy-Mobile condition. \textbf{Panel A}: MTL, RTL and Surprise reward. \textbf{Panel B}: MTL and RTL combined with Surprise. \textbf{Panel C}: CAIS and CAIS + Surprise reward. \textbf{Panel D}: Resultant Q values of the $MOVE$ action of different rewards combined with Surprise. Shaded area suggest 1 standard deviation.}
    \label{fig4}
\end{figure}

The central and most critical findings emerge from the "Noisy-Mobile" condition, which introduces a significant confounding variable: a randomly-directed external force of 40N applied to the mobile at every physics update. This condition is designed to directly test the hypothesis that robust agency detection requires explicit causal inference, not mere correlation. In this noisy setting, the RTL and MTL reward signals become completely uninformative. As shown in Figure \ref{fig4} Panel A, the reward signal is high and erratic throughout all phases, with no discernible difference between baseline, attached, and extinction periods. The agent's own influence is "covered in the noisy movement," and the signal from its actions is drowned in the noise of the external force. The RL agent, unable to distinguish between movement it caused and random environmental fluctuations, reinforces spurious correlations and fails to learn.\par

In stark contrast, the CAIS mechanism continues to function robustly. As depicted in Figure \ref{fig4}, Panel C, the CAIS reward for the attached limb shows a distinct and sustained increase during the attached phase, cleanly separating it from the baseline reward of the unattached limbs. CAIS succeeds because it is not concerned with the absolute magnitude of movement but with the change in the statistical properties of the movement contingent on a specific action. It compares the conditional outcome distribution $p(h|a)$ with the unconditional baseline distribution $p(h)$. The natural outcome is captured in the baseline distribution $p(h)$. The contingency of the attached limb systematically shifts the outcome distribution $p(h|a_{attached})$ relative to this baseline. The 1-Wasserstein distance, which CAIS uses, is highly sensitive to this shift, generating a high reward and allowing the agent to filter out the unstructured environmental noise to isolate its specific causal influence. Consequently, as shown by the learned Q-values in Figure \ref{fig4}, Panel C, only the agent guided by CAIS learns the correct policy in the noisy environment, providing powerful evidence that robust agency requires mechanisms that move beyond simple correlation to explicitly infer causal influence.\par

\subsection{Modeling Expectation Violation and the Extinction Burst}
Having established the robustness of CAIS, the analysis turns to the model's psychological plausibility by testing its ability to reproduce the "extinction burst"—an intensified, temporary increase in a previously reinforced behavior when the reinforcement is unexpectedly removed. This is modeled by augmenting the primary reward with a "Surprise" signal, defined as the prediction error between the agent's expectation of a sensory outcome, $E[h|a]$, and the actual outcome observed, $h_{t+1}$.\par

The surprise term effectively tracks violations of the agent's learned expectations. As shown in Figure \ref{fig3}, Panel A, the surprise signal spikes at the beginning of the attached phase, signaling the new contingency, and then decreases as the agent learns to predict it. Crucially, it spikes dramatically again at the onset of the extinction phase, when the expected movement fails to occur. This demonstrates that the surprise signal functions as an effective expectation-violation detector.\par

When this surprise signal is added to the CAIS reward, the agent exhibits behavior highly consistent with the extinction burst phenomenon. As shown in Figure \ref{fig3}, Panel C for the Free-Mobile condition and, more critically, in Figure \ref{fig4}, Panel C for the Noisy-Mobile condition, the combined reward shows a noticeable spike at the beginning of the extinction phase. This reward spike provides a strong learning signal that drives a temporary increase in the agent's policy to kick the (now ineffective) limb, successfully replicating the burst behavior. This success is a direct result of the synergy between the high-quality predictive world model required for CAIS and the surprise mechanism. Because the CAIS model learned the true contingency by filtering out noise, its expectation is strong and precise. The violation of this strong expectation generates a clean and reliable surprise signal, which in turn drives a coherent behavioral response.\par

In contrast, Panel B in both Figures \ref{fig3} and \ref{fig4} shows that Surprise does not produce a clear extinction burst when combined with RTL or MTL. In the noisy condition, the base reward signal is already high and erratic, masking the effect of the surprise term. In the clean condition, the large drop in the base reward at the onset of extinction overshadows the positive surprise signal, preventing a net increase in reward that would drive a behavioral burst.\par

\subsection{Comparative Analysis and Summary of Learned Policies}
To synthesize these findings, this section provides a direct comparison of the outcomes for each reward strategy across both environmental conditions. The agent's final learned policy—its propensity to act, as reflected in its Q-values—serves as the ultimate measure of success. The results unequivocally demonstrate the superiority of the causal inference approach for developing a robust sense of agency.\par

\begin{itemize}
    \item \textbf{RTL-based Agent:} This agent represents a purely correlational, perceptual strategy. It learns the correct policy in the simple, deterministic Free-Mobile environment but fails completely in the Noisy-Mobile environment. Its learned sense of agency is brittle, an artifact of a clean environment, and does not generalize to more ecologically realistic settings with confounding factors.
    \item \textbf{CAIS-based Agent:} This agent represents a true causal inference strategy. It successfully learns the correct policy in both the simple and noisy environments. Its ability to compare outcome distributions allows it to filter out confounding noise and identify its true causal power, resulting in a robust and generalizable sense of agency.
    \item \textbf{CAIS + Surprise Agent:} This agent builds upon the robust foundation of CAIS to incorporate a psychologically-inspired mechanism of expectation violation. It not only learns robustly in both environments but also successfully replicates the extinction burst phenomenon. As noted in the initial analysis, this combination is the only one that "clearly shows the desired behavior: the increased moving probability of attached limb in the attached phase, and an extinction burst in the extinction phase" in the challenging noisy condition. This makes it the most complete and plausible model of contingency learning presented in this work.
\end{itemize}

\section{Conclusion}
\label{sec:conclusion}

This research provides strong computational evidence that robust agency detection requires a shift from correlational to causal-inferential methods, particularly in complex environments. The Causal Action Influence Score (CAIS) succeeded where perceptual rewards failed because it is designed to explicitly measure the statistical divergence an action creates, allowing it to filter out confounding environmental noise and isolate the agent's true efficacy. This robust causal model proved to be a prerequisite for modeling more complex cognitive phenomena. The successful simulation of the "extinction burst" validates this, as a meaningful surprise signal can only be generated when a strong, precise expectation is violated—an expectation that the noise-muddled, correlation-based agent could not form. The technical robustness of CAIS stems from its novel use of the 1-Wasserstein distance to compare sensory outcome distributions, a method well-suited to detecting an action's systematic influence amidst noise.\par

These findings can be situated within developmental psychology. The agent's learning process serves as a computational analogue for an infant's discovery of the "ecological self"—the self as an agent perceived through direct environmental interaction. Developmental theory suggests a "contingency switch" where an infant's preference shifts from the perfect contingency of self-interaction to the high-but-imperfect contingency of social partners. Our model successfully simulates the critical pre-switch phase, building the foundational sense of agency. This distinction has direct implications for Human-Robot Interaction (HRI), where perfect, machine-like contingency can feel un-social. The CAIS framework, by providing a graded measure of influence, offers a tool to design robots that can target an optimal, imperfect level of responsiveness, fostering more natural engagement.\par

The study's simplifications—a constrained agent morphology, discrete action space, and pre-trained visual encoder—were necessary to isolate the core research question but also highlight clear paths for future work. The next steps include computationally modeling the "contingency switch" itself, likely by introducing probabilistic action-outcome links, and validating the agent's robustness by transferring it to a physical robotic platform. Integrating the CAIS mechanism into agents that learn representations and continuous actions end-to-end remains a key long-term objective.\par

In conclusion, this work demonstrates that by moving beyond brittle correlations and explicitly formalizing agency detection as a problem of causal inference, an artificial agent can develop a robust and generalizable sense of its own efficacy. The Causal Action Influence Score provides a novel and powerful intrinsic reward that enables learning even in the presence of significant environmental confounding, where traditional perceptual methods fail. By situating these findings within the rich theoretical landscapes of developmental psychology and human-robot interaction, this research not only provides a psychologically plausible model of a foundational cognitive process but also offers a practical framework for building more adaptive and capable autonomous systems. The principles elucidated here—of isolating causal influence, of building robust expectations, and of understanding the nuanced role of contingency in interaction—represent a concrete and promising step toward creating artificial agents that can discover their own capabilities and function effectively in the rich uncertainty of the real world.\par

\bibliographystyle{unsrt}  
\bibliography{references}

\newpage

\section*{Appendix}
\label{sec:appendix}

\subsection*{Agent Architecture}
\label{subsec:agent}
\begin{figure}[h]
    \centering
    \includegraphics[width=0.5\textwidth]{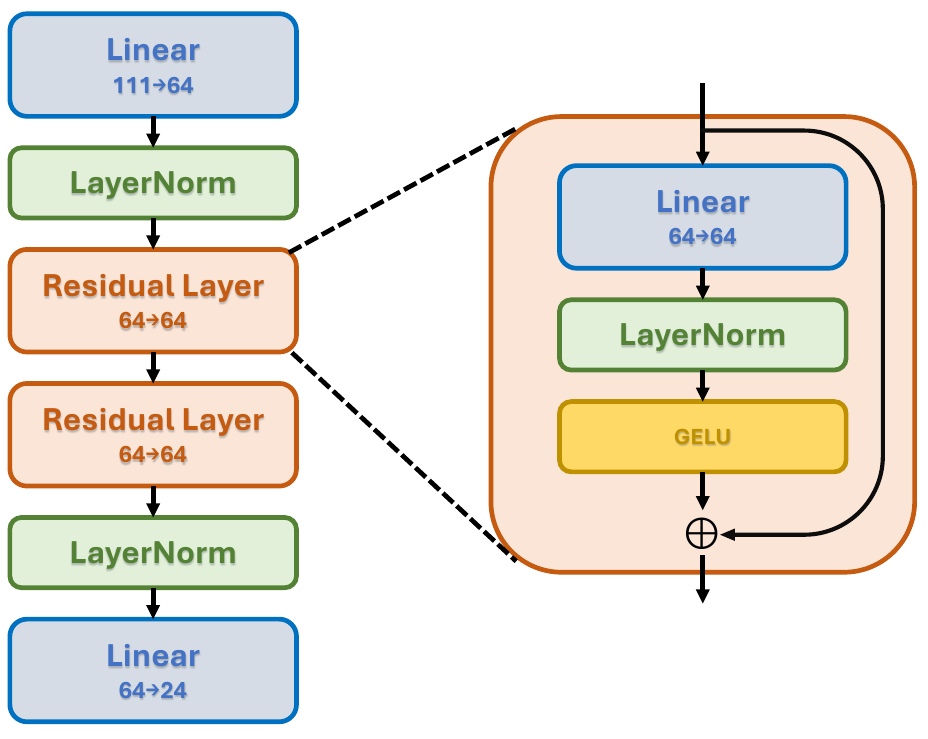}
    \caption{Agent Structure}
    \label{fig5}
\end{figure}

The SARSA agent is implemented as a neural network policy, with its architecture detailed in Figure \ref{fig5}.

\begin{enumerate}
    \item \textbf{Input Processing:} The network begins by concatenating the agent's proprioceptive state (47 dimensions) and its visual representation (64 dimensions). This combined input vector is fed into a linear layer, which compresses it into a 64-dimensional hidden state. This state is then normalized using Layer Normalization \cite{ba_layer_2016}.
    \item \textbf{Core Processing:} The normalized representation is passed through two sequential residual layers. Each residual layer is composed of a linear layer, Layer Normalization, a GELU activation, and a skip connection.
    \item \textbf{Output Head:} Finally, the output from the residual block undergoes another Layer Normalization. It is then transformed by a final linear layer to produce the desired output, a 24-dimensional vector representing two control values for each of MIMo's 12 joints.
\end{enumerate}

\subsection*{MIMo's proprioceptive sensors and actuators}
We implemented 12 actuators in MIMo's four limbs, comprising horizontal and abduction rotators for the shoulders, flexion and abduction rotators for the hips, and hinge actuators for the elbows and knees. For its proprioceptive state input, the agent has access to the angles of all joints. However, its actions are exclusively applied to the specified actuators. Table \ref{tab:mimo_proprio} contains a comprehensive list of all joints and indicates which are actuated.

\begin{table}[h]
    \centering
    \begin{tabular}{|c|c|c|c|c|c|c|c|}
        \hline 
        hip\_lean1 & hip\_lean2 & hip\_rot1 & hip\_rot2 & hip\_bend1 & hip\_bend2\tabularnewline
        \hline 
        head1 & head2 & head3 & left\_eye1 & left\_eye2 & left\_eye3\tabularnewline
        \hline 
        right\_eye1 & right\_eye2 & right\_eye3 & \textbf{right\_shoulder1} & \textbf{right\_shoulder2} & right\_shoulder3\tabularnewline
        \hline 
        \textbf{left\_shoulder1} & \textbf{left\_shoulder2} & left\_shoulder3 & \textbf{right\_elbow} & \textbf{left\_elbow} & right\_hand1\tabularnewline
        \hline 
        right\_hand2 & right\_hand3 & left\_hand1 & left\_hand2 & left\_hand3 & right\_fingers\tabularnewline
        \hline 
        left\_fingers & \textbf{right\_hip1} & \textbf{right\_hip2} & right\_hip3 & \textbf{left\_hip1} & \textbf{left\_hip2}\tabularnewline
        \hline 
        left\_hip3 & \textbf{right\_knee} & \textbf{left\_knee} & right\_foot1 & right\_foot2 & right\_foot3\tabularnewline
        \hline 
        left\_foot1 & left\_foot2 & left\_foot3 & right\_toes & left\_toes & \tabularnewline
        \hline 
    \end{tabular}
    \caption{MIMo's joint names. Ball joints are implemented as 3 hinge joints. The hip joint has specifically 2 higne joints for each of the 3 dimensions of the hip. Bold names mean actuated joints.}
    \label{tab:mimo_proprio}
\end{table}

\subsection*{Visual encoder}
\label{appendix:encoder}
Our U-Net model features a symmetrical design with an encoder of four down-sampling layers and a decoder of four up-sampling layers. Each layer consists of a convolution, an Instance Normalization \cite{ulyanov_instance_2017}, and a GELU activation function \cite{hendrycks_gaussian_2023}, followed by a sampling operator.\par

The output from the final down-sampling layer (the bottleneck) serves as the core feature representations. These representations are used in two ways:
\begin{enumerate}
    \item \textbf{Image Reconstruction:} They are fed through the up-sampling layers, incorporating skip-connections from the corresponding encoder layers, to reconstruct the input image. A reconstruction loss is applied to this final output.
    \item \textbf{Temporal Learning:} The representations are also passed to a projector, which maps them to a 512-dimensional latent space. We then apply the CLTT loss in this space to ensure the representations encode rich temporal information.
\end{enumerate}

The CLTT loss is then applied in this space: 
\begin{equation}
\mathcal{L}\left(z_{i}\right)=-\log\frac{\exp\left({\rm sim}\left(z_{i},z'_{i}\right)/\tau\right)}{\sum_{k=1,k\neq i}^{N_{B}}\left[\exp\left({\rm sim}\left(z_{i},z_{k}\right)\right)+\exp\left({\rm sim}\left(z_{i},z'_{k}\right)\right)\right]/\tau},
\label{eq:cltt}
\end{equation}
where $z_{i}$ and $z'_{i}$ are the projected latent codes of an adjacent pair of visual input images, and $N_{B}$ is the number of pairs in a batch. We use a batch size of $N_B=50$ in pretraining. Specifically, we adopt the cosine similarity as our similarity function: ${\rm sim}\left(u,v\right)=\frac{u^{\top}v}{\left\Vert u\right\Vert \left\Vert v\right\Vert }$. The softmax temperature $\tau$ from the original SimCLR loss is set to $\tau=0.5$.
The CLTT loss ensures the representations capture rich temporal dynamics.

\end{document}